\PassOptionsToPackage{table}{xcolor}
\documentclass[twocolumn,letterpaper]{article} 
\usepackage{aaai2026}  
\usepackage{times}  
\usepackage{helvet}  
\usepackage{courier}  
\usepackage[hyphens]{url}  
\usepackage{graphicx} 
\usepackage{natbib}  
\usepackage{caption} 
\usepackage{algorithm}
\usepackage{algorithmic}
\usepackage{booktabs} 
\usepackage{multirow}
\usepackage{amsmath}
\usepackage{amsfonts}
\usepackage{array} 
\usepackage{dsfont}
\usepackage{newfloat}
\usepackage{listings}
\newcolumntype{L}[1]{>{\raggedright\arraybackslash}p{#1}}

\DeclareCaptionStyle{ruled}{labelfont=normalfont,labelsep=colon,strut=off} 
\floatstyle{ruled}
\newfloat{listing}{tb}{lst}{}
\floatname{listing}{Listing}
\nocopyright 

\title{LeAdQA: LLM-Driven Context-Aware Temporal Grounding for Video Question Answering}
\author{
    Xinxin Dong\textsuperscript{\rm 1},
    Baoyun Peng\textsuperscript{\rm 2}\thanks{Corresponding Author.},
    Haokai Ma\textsuperscript{\rm 3},\\
    Yufei Wang\textsuperscript{\rm 1},
    Zixuan Dong\textsuperscript{\rm 1},
    Fei Hu\textsuperscript{\rm 1},
    Xiaodong Wang\textsuperscript{\rm 1},
}
\affiliations{
    \textsuperscript{\rm 1}National Key Laboratory of Parallel and Distributed Computing, College of Computer Science and Technology, \\ National University of Defense Technology\\

    \textsuperscript{\rm 2}Academy of Military Sciences\\
    \textsuperscript{\rm 3}National University of Singapore\\
pengbaoyun13@alumni.nudt.edu.cn
}

\usepackage{bibentry}

\begin{document}

\maketitle

\begin{abstract}

Video Question Answering (VideoQA) requires identifying critical moments in long videos and reasoning about their causal relationships to answer semantically complex questions. Current approaches suffer from two key limitations: (1) task-agnostic sampling that overwhelms relevant content with irrelevance, and (2) heuristic retrieval that captures superficial patterns while missing causal-temporal structures essential for complex reasoning. To tackle these limitations, we propose LeAdQA, a novel approach that combines causal-aware query refinement with fine-grained visual grounding. Our method leverages LLMs to reformulate question-option pairs, resolving causal ambiguities and sharpening temporal focus. These refined queries guide precise retrieval of salient segments, while an adaptive fusion mechanism integrates evidence to maximize relevance. Finally, an MLLM processes the integrated visual-textual cues to generate contextually-grounded answers. LeAdQA achieves state-of-the-art performance on the NExT-QA, IntentQA, and NExT-GQA datasets, demonstrating that precise visual clues significantly enhances the model's reasoning on complex questions while maintaining computational efficiency.

\end{abstract}

\section{Introduction}

Video understanding is increasingly critical across diverse domains, spanning educational platforms, entertainment systems, surveillance networks, and autonomous vehicles~\cite{wang2024videotree,maaz2023video}. Within this landscape, Video Question Answering (VideoQA) emerges as a foundational capability that requires models to comprehend complex spatio-temporal dynamics and reason about visual content in response to queries in natural language~\cite{xiao2021next,li2023intentqa}. 
However, transitioning from short-form to long-form video analysis introduces unprecedented challenges that reshape the problem space. Extended video sequences exhibit a sparse distribution of critical information within redundant content, posing two core technical challenges: (1) identifying and extracting relevant visual cues across temporally distant segments; (2) maintaining fine-grained temporal resolution while processing computationally intensive long sequences.

 \begin{figure}[!t]
    \centering
    \includegraphics[width=1\linewidth]{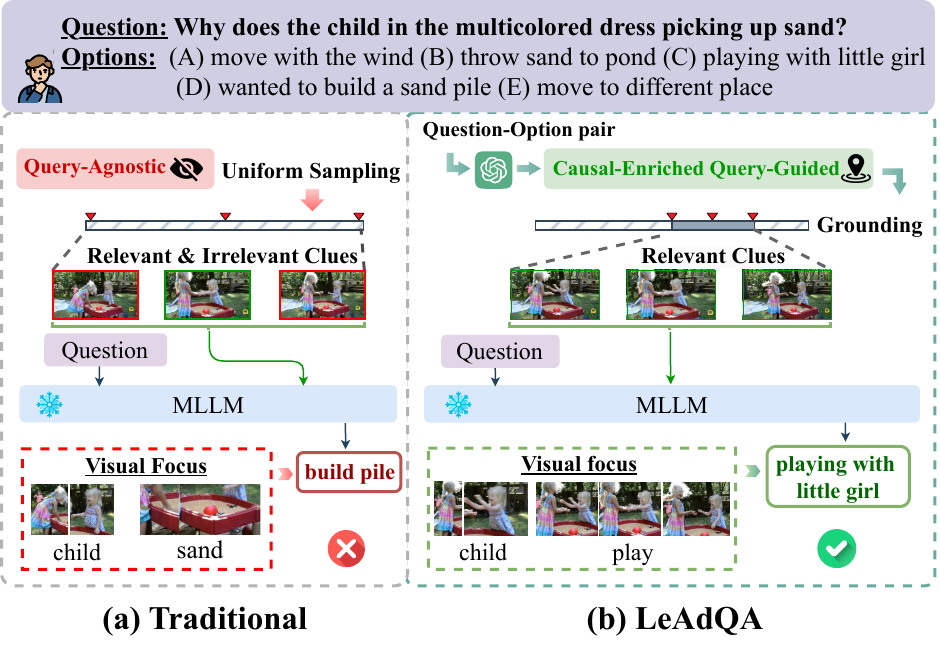}
    \caption{Architecture comparison: (a) Traditional frameworks incorporate irrelevant spatiotemporal data, hindering visual reasoning; (b) LeAdQA enables precision localization of query-relevant moments via temporal grounding.
    }
    \label{fig: figure 1}
\end{figure}

Early VideoQA methods relied on 3D convolutional networks~\cite{tran2015learning}, hierarchical modeling~\cite{lu2016hierarchical}, and attention-based frame localization~\cite{ren2016faster} to capture spatiotemporal patterns. However, these approaches primarily focused on surface-level correlations and lacked the semantic abstraction required for complex reasoning~\cite{lei2018tvqa}.
The emergence of Multimodal Large Language Models (MLLMs) has revolutionized video understanding~\cite{liu2024visual,wang2407tarsier}. 
Current MLLM-based approaches fall into three paradigms: end-to-end methods processing visual and textual tokens jointly~\cite{wang2407tarsier,maaz2023video}, two-stage approaches separating visual understanding from linguistic reasoning~\cite{zhang2023simple,yu2024self}, and hybrid frameworks combining both strategies~\cite{wang2024videotree,li2023intentqa}. These MLLM-based methods have achieved remarkable progress in video question answering by leveraging the inherent reasoning capabilities of large language models (LLMs) to bridge the semantic gap between visual content and natural language queries, enabling more sophisticated inference beyond traditional pattern matching approaches.

Despite these advances, current VideoQA methods remain constrained in long-form understanding due to several limitations. Foremost, the neglect of causal relationships between questions and candidate answers leads to treating options as isolated entities~\cite{wei2023visual,zang2023discovering}—overlooking semantic interdependencies that could enable comparative reasoning critical for multiple-choice scenarios.
Such cognitive gap propagates to visual processing, where coarse temporal grounding induces fragmented localization~\cite{guo2024trace,wu2025number}: critical moments splinter into misaligned segments that fracture causal chains (e.g., separating triggers from consequences).
Compounding these errors, the inherent redundancy of long videos forces models into a lose-lose computational bind~\cite{wang2024retake}—where global processing may drown signals in noise, while aggressive sampling discards sparse pivotal events.

To address these challenges, we present \textbf{LeAdQA}, a novel \textbf{L}LM-Driv\textbf{e}n Context-\textbf{A}ware Temporal Groun\textbf{d}ing framework that enhances MLLMs for VideoQA through integrated causal-temporal reasoning. Our approach operates through three key innovations:
First, LeAdQA leverages LLMs to reformulate question-option (Q-O) pairs, explicitly injecting causal relationships to resolve linguistic ambiguities and establishing "leading" contextual cues for reasoning. 
Second, unlike traditional frameworks that incorporate irrelevant spatiotemporal data, LeAdQA enables precision localization of query-relevant moments through dedicated temporal grounding, as illustrated in Figure~\ref{fig: figure 1}.
These causally-enhanced queries drive a cross-modal transformer that dynamically localizes critical video segments by aligning textual semantics with visual content, significantly outperforming question-only localization methods. 
Third, we introduce an adaptive interval fusion mechanism that evaluates candidate segments using dual criteria: temporal overlap (IoU) and salience scores. This filtering approach preserves semantically coherent intervals while eliminating noise, enabling MLLMs to effectively model Q-O causality alongside visual features for precise temporal reasoning in long-form VideoQA.

Extensive experiments on NExT-QA~\cite{xiao2021next}, IntentQA~\cite{li2023intentqa}, and NExT-GQA~\cite {xiao2024can} demonstrate consistent improvements over state-of-the-art (SOTA) approaches. Our empirical analysis reveals that LLM-mediated causal resolution enhances temporal grounding precision, with higher tIoU strongly correlating with QA accuracy, confirming that informational quality is more critical than quantity.
Our main contributions include: 
 \begin{itemize} 
    \item We present LeAdQA, a novel LLM-driven architecture that addresses causal-temporal gaps through: rewriting question-option pairs to inject causal dependencies, leveraging option semantics as cross-modal constraints, and generating refined temporal proposals via context-aware grounding.
  
     \item We propose an \textbf{adaptive NMS module} that combines temporal overlap with LLM-extracted causal relevance scores, preserving coherent segments while eliminating redundancy.
    \item Comprehensive evaluation across three datasets demonstrates significant improvements in temporal localization and QA accuracy.
\end{itemize} 

\section{Related Works}
\subsection{Video Temporal Grounding}

Video Temporal Grounding (VTG) localizes moments in untrimmed videos that semantically align with textual queries. Current approaches follow two paradigms: two-stage and end-to-end methods. Two-stage approaches generate temporal proposals~\cite{gao2017tall,xu2018text} then perform cross-modal matching, but suffer from computational inefficiency due to dense candidate sampling. End-to-end methods directly regress temporal boundaries, with transformer-based approaches like MomentDETR~\cite{lei2021detecting} and QD-DETR~\cite{moon2023query} formulating VTG as set prediction with enhanced cross-attention. Despite efficiency gains, current methods still struggle with long-range dependencies and precise alignment. Unlike VTG's descriptive event-boundary queries, VideoQA grounding requires multimodal reasoning for discriminative answer selection.

\subsection{Multimodal Large Language Models}
The success of large language models (LLMs) in natural language processing has sparked increasing interest in extending their capabilities to multimodal tasks~\cite{radford2021clip, liu2024visual}. One line of research translates non-textual inputs into natural language to align visual and textual modalities. For instance, OFA~\cite{wang2022ofa} performs visual-to-text translation while LaViLa~\cite{zhao2023learning} generates video captions. Alternatively, some approaches employ trainable interface layers for direct modality bridging. Flamingo~\cite{alayrac2022flamingo} links CLIP~\cite{radford2021clip} vision encoders to LLMs via learned projections, while BLIP-2~\cite{liBLIP2BootstrappingLanguageImage2023} and LLaVA~\cite{liu2024visual} further refine alignment using Q-Former and MLP-based adapters, respectively. Recent video-oriented models such as Video-ChatGPT~\cite{maaz2023video} incorporate temporal dynamics to support video-language understanding. Although effective, these methods often rely on intensive cross-modal training and incur significant computational overhead. 

\subsection{Video Question Answering}
VideoQA aims to answer questions based on video content and textual queries, posing challenges in spatiotemporal reasoning and cross-modal alignment. Early approaches adopted cross-attention mechanisms~\cite{chu2018forgettable} for visual-textual alignment. Subsequent work explores memory networks~\cite{yu2020long} for multi-hop reasoning and graph neural networks~\cite{seo2021attend} to model object-scene interactions. More recent advances leverage pre-trained vision-language models and LLMs, with strategies such as fine-tuning~\cite{tan2024koala}, constructing spatial-temporal scene graphs~\cite{fei2024video}, and filtering captions without additional training~\cite{zhang2023simple,islam2024video}. VideoAgent~\cite{wang2025videoagent} futher employs LLMs as iterative information extractors. While effective, many of these methods underutilize fine-grained visual cues. In contrast, our framework can better capture detailed visual semantics to improve complex reasoning.

\begin{figure*}[!htbp]
    \centering
    \includegraphics[width=0.95\textwidth]{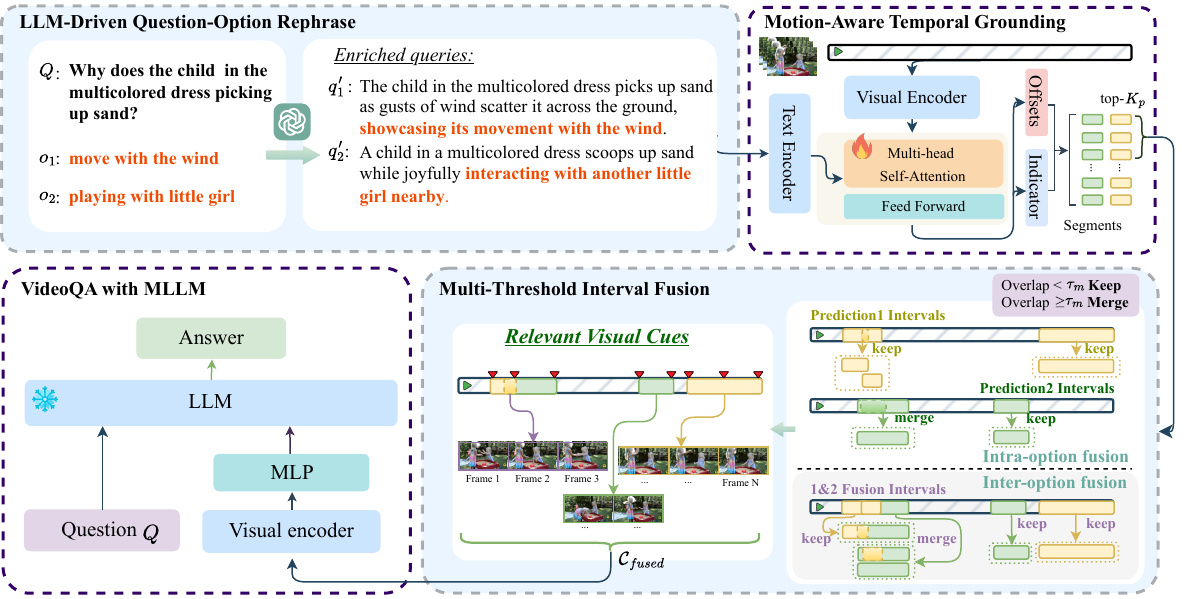}
    \caption{The architecture of LeAdQA. Question-option pairs are first rephrased by LLMs to generate enhanced descriptions, which are then used to localize relevant visual segments. Temporal intervals are subsequently filtered and merged through overlap threshold analysis. Finally, the optimized segments are fed into an MLLM to generate the final answer.}

    \label{fig: framework}
\end{figure*}

\section{Method}

We present \textbf{LeAdQA}, a novel VideoQA framework that integrates causal-aware language modeling and temporal grounding to enhance MLLMs in contextual understanding and answer reasoning. As illustrated in Figure~\ref{fig: framework}, LeAdQA comprises three core components: (1) LLM-guided question-option rephrasing, (2) motion-aware temporal grounding, and (3) multi-threshold interval fusion for robust localization.

\subsection{Problem Formulation}

Given a video $\mathcal{V}$ and a multiple-choice question $Q$ with answer candidates $\mathcal{O} = \{o_i\}_{i=1}^N$, where $N$ is the number of candidate options, we define $q_i = (Q, o_i)$ as the query corresponding to option $o_i$. LeAdQA aims to (i) generate semantically enriched query descriptions $\{q_i'\}_{i=1}^N$ using an LLM, (ii) localize temporally grounded video segments that support or contradict each $q_i'$, and (iii) aggregate temporal evidence for answer prediction.

\subsection{LLM-Driven Question-Option Rephrase}

To enhance causal reasoning and video alignment, each query $q_i'$ is rephrased using an LLM $\mathcal{M}_r$ with a structured prompt $\mathcal{P}_r$:

\begin{equation}
q_i' = \mathcal{M}_r(\mathcal{P}_r(Q, o_i)).
\end{equation}

The output $q_i'$ describes a hypothetical video situation under which option $o_i$ is true, making implicit causal cues explicit. The full set of enriched queries is denoted as $\mathcal{Q}' = \{q_i'\}_{i=1}^N$.

\subsection{Motion-Aware Temporal Grounding}
\subsubsection{Unified Formulation.}
Following the unified spatiotemporal alignment framework~\cite{lin2023univtg}, we divide video $\mathcal{V}$ into a sequence of overlapping clips $\{v_j\}_{j=1}^{L_v}$ of fixed duration $l$, centered at timestamps ${t_j}$. For each clip $v_j$ and query $q_i'$, the model predicts:

\begin{itemize}
  \item \textbf{Foreground Flag:} $f_{ij} \in \{0, 1\}$: Indicates whether $v_j$ is relevant to query $q_i'$.

  \item \textbf{Boundary Offset:} $\delta_{ij} = [\delta_{ij}^s, \delta_{ij}^e]$: Denotes the temporal offset from $t_j$ to the predicted segment boundaries. 

  \item \textbf{Saliency Score:} $s_{ij} \in [0, 1]$: Measures semantic relevance between $v_j$ and query $q_i'$.
\end{itemize}

The resulting grounded segment is:
\begin{equation}
b_{ij} = [t_j - \delta_{ij}^s, t_j + \delta_{ij}^e],
\end{equation}
and the set of relevant segments for query $q_i'$ is $\mathcal{B}_i = \{b_{ij} \mid f_{ij} = 1 \land s_{ij} > \tau_p\}$. Here, $\tau_p \in [0, 1]$ is a saliency threshold that filters out low-relevance segments. It is tuned on a validation set and held fixed during inference.

\subsubsection{Architecture and Training}

We encode an enriched query $q_i'$ with $L_q$ tokens and a video $\mathcal{V}$ with $L_v$ clips separately using text and video encoders, followed by FFNs to project features into a shared $d$-dimensional space, obtaining token-level features $\mathbf{F}_t$ and clip-level features $\mathbf{F}_v$ respectively:

\begin{align}
\mathbf{F}_t &= \{q_{i',j}\}_{j=1}^{L_q} \in \mathbb{R}^{L_q \times d} \\
\mathbf{F}_v &= \{v_j\}_{j=1}^{L_v} \in \mathbb{R}^{L_v \times d} 
\end{align}

\paragraph{Cross-Modal Fusion.}  We apply modality and positional embeddings to get $\tilde{\mathbf{F}}_v$ and $\tilde{\mathbf{F}}_t$, concatenate them into $\mathbf{F}_z = [\tilde{\mathbf{F}}_v; \tilde{\mathbf{F}}_t]$, and are processed by $K_t$ Transformer layers before being passed to the subsequent prediction head:

\begin{equation}
\mathbf{F}_z^{(l)} = \text{MLP}(\text{MSA}(\mathbf{F}_z^{(l-1)})), \quad l = 1, \dots, K_t.
\end{equation}

\paragraph{Prediction Heads and Losses.} We employ three prediction heads to optimize distinct grounding objectives:
\begin{itemize}
\item \textbf{Foreground Head}: This head predicts the binary label $\hat{f}_{ij}$ for each clip $v_j$ using a stack of $1 \times 3$ convolutional layers with ReLU activation. The training objective is binary cross-entropy:
\begin{equation}
\mathcal{L}_{fg} = -\sum_{j=1}^{L_v} [f_{ij} \log \hat{f}_{ij} + (1 - f_{ij}) \log (1 - \hat{f}_{ij})]
\end{equation}

\item \textbf{Boundary Head}: This head predicts the segment boundaries $\hat{\delta}_{ij}^s$ and $\hat{\delta}_{ij}^e$ for each foreground clip. The loss combines Smooth-$L_1$ distance between predicted and ground-truth offsets, and an IoU-based loss between predicted segment $\hat{b}_{ij}$ and ground-truth $b_{ij}$:
\begin{equation}
\mathcal{L}_{reg} = \mathds{1}_{\{f_{ij} = 1\}} [\lambda_{L1} \mathcal{L}_{\mathrm{smooth}} + \lambda_{\mathrm{IoU}} \mathcal{L}_{\mathrm{IoU}}]
\end{equation}
\item \textbf{Saliency Head}: This head computes the cosine similarity between visual and textual embeddings to assign a saliency score $\hat{s}_{ij}$. We incorporate both intra-video and inter-video contrastive learning to encourage fine-grained and global discriminability:
\begin{equation}
\mathcal{L}_{sal} = \lambda_{\mathrm{inter}} \mathcal{L}_{sal}^{inter} + \lambda_{\mathrm{intra}} \mathcal{L}_{sal}^{intra}
\end{equation}
\end{itemize}

The total loss over all clips and all query-option pairs is:

\begin{equation}
\mathcal{L}_{total} = \frac{1}{N} \sum_{i=1}^N \sum_{j=1}^{L_v} (\mathcal{L}_{fg} + \mathcal{L}_{reg} + \mathcal{L}_{sal})
\end{equation}

\subsection{Multi-Threshold Interval Fusion}

After grounding, we obtain $K_s$ top-ranked segments for each query $q_i'$, resulting in the candidate set $\mathcal{C} = \{c_k = [t_k^s, t_k^e]\}_{k=1}^{N \times K_s}$. These segments represent the model's predicted intervals, with $t_k^s$ and $t_k^e$ denoting the start and end timestamps, respectively. The segments are then refined through merging and fusion to reduce redundancy and enhance temporal grounding.

\paragraph{IoU-Based Merging.} 
Two temporal intervals $c_k$ and $c_{k'}$ are merged if their Intersection over Union (IoU) exceeds a predefined threshold $\tau_m$:
\begin{equation}
    \mathrm{IoU}(c_k, c_{k'}) = \frac{\mathrm{overlap}}{\mathrm{union}} \geq \tau_m
\end{equation}

\noindent where:
\begin{align}
    \mathrm{overlap} &= \max(0, \min(t_k^e, t_{k'}^e) - \max(t_k^s, t_{k'}^s)) \\
    \mathrm{union} &= (t_k^e - t_k^s) + (t_{k'}^e - t_{k'}^s) - \mathrm{overlap}
\end{align}
If IoU exceeds $\tau_m$, the intervals are merged:
\begin{equation}
    \mathrm{Merge}(c_k, c_{k'}) = [\min(t_k^s, t_{k'}^s), \max(t_k^e, t_{k'}^e)].
\end{equation}
This approach reduces temporal redundancy by retaining only the most relevant intervals.

\paragraph{Hierarchical Fusion Strategy.} 
We employ a two-stage hierarchical fusion process to integrate temporal evidence:
\begin{itemize}
    \item \textbf{Intra-option fusion}: Consolidates overlapping temporal intervals within each enriched query to eliminate redundant segments and create coherent evidence units.
    \item \textbf{Inter-option fusion}: Aggregates the consolidated intervals across different answer options to capture contextual information and enable comparative reasoning.
\end{itemize}

These fusion steps help to maintain a diverse and non-redundant set of segments, improving the accuracy of temporal grounding.

During inference, all prediction heads contribute, and NMS is applied to remove redundant intervals based on high overlap. The final non-redundant segments, $\mathcal{C}_{\mathrm{fused}}$, are used for answer selection, ensuring that the most relevant video segments are chosen to answer the query.

\begin{table*}[!htbp]
	\centering
	\setlength{\tabcolsep}{4pt} 
	\begin{tabular*}{\textwidth}{@{} p{5.5cm} c p{1.5cm} *{6}{c} @{}}
			
		\toprule
		\multirow{2}{*}{\textbf{Methods}} & \textbf{Answering} & \multicolumn{7}{c}{\textbf{Grounding}} \\
		\cmidrule(l){2-2}
		\cmidrule(lr){3-9}
		~ & Acc@QA & Acc@GQA & mIoP & IoP@0.3 & IoP@0.5 & mIoU & IoU@0.3 & IoU@0.5 \\
		\midrule

		\cellcolor{gray!30}Human  &
		\cellcolor{gray!30}93.3 & \cellcolor{gray!30}82.1 &
		\cellcolor{gray!30}72.1 & \cellcolor{gray!30}91.7 &
		\cellcolor{gray!30}86.2 & \cellcolor{gray!30}61.2 &
		\cellcolor{gray!30}86.9 & \cellcolor{gray!30}70.3 \\
		
		\cellcolor{gray!30}Random &
		\cellcolor{gray!30}20.0 & \cellcolor{gray!30}1.7 &
		\cellcolor{gray!30}21.1 & \cellcolor{gray!30}20.6 &
		\cellcolor{gray!30}8.7  & \cellcolor{gray!30}21.1 &
		\cellcolor{gray!30}20.6 & \cellcolor{gray!30}8.7 \\
		\midrule
		
		IGV~\cite{li2022invariant}      & 50.1 & 10.2 & 21.4 & 26.9 & 18.9 & 14.0 & 19.8 & 9.6 \\
		VGT~\cite{xiao2024can}          & 50.9 & 12.7 & 24.7 & 26.0 & 24.6 & 3.0  & 4.2  & 1.4 \\
		VIOLETv2~\cite{fu2023empirical} & 52.9 & 12.8 & 23.6 & 25.1 & 23.3 & 3.0  & 4.3  & 1.3 \\
		Temp [Swin]~\cite{xiao2024can}  & 55.9 & 14.4 & 25.3 & 26.4 & 25.3 & 3.0  & 3.6  & 1.7 \\
		Temp [CLIP]~\cite{xiao2024can}  & 59.4 & 14.7 & 24.1 & 26.2 & 24.1 & 6.8  & 8.3  & 3.7 \\
		Temp [CLIP] (NG+)~\cite{xiao2024can} & 60.2 & 16.0 & 25.7 & 31.4 & 25.5 & 12.1 & 17.5 & 8.9 \\
		FrozenBiLM~\cite{yang2022zero}  & 69.1 & 15.8 & 22.7 & 25.8 & 22.1 & 7.1  & 10.0 & 4.4 \\
		FrozenBiLM (NG+)~\cite{xiao2024can} & \underline{70.8} & 17.5 & 24.2 & 28.5 & 23.7 & 9.6 & 13.5 & 6.1 \\
		SeViLA~\cite{yu2024self}        & 68.1 & 16.6 & \underline{29.5} & \underline{34.7} & 22.9 & \textbf{21.7} & \underline{29.2} & \underline{13.8} \\
		QGAC-TR~\cite{ExploringQuestionGuidance} & 63.6 & \underline{18.3} & 28.3 & 32.8 & \underline{27.7} & 15.7 & 18.6 & 11.7 \\
		\midrule
		\textbf{LeAdQA-7B}  & 67.5 & 19.2 & \multirow{2}{*}{\textbf{30.3}} & \multirow{2}{*}{\textbf{39.6}} & \multirow{2}{*}{\textbf{29.5}} & \multirow{2}{*}{\underline{20.5}} & \multirow{2}{*}{\textbf{31.3}} & \multirow{2}{*}{\textbf{17.5}} \\
		\textbf{LeAdQA-34B} & \textbf{77.0} & \textbf{20.7} &  &  &  &  &  &  \\
		\bottomrule
	\end{tabular*}
	\caption{Performance on NExT-GQA test set. Answering and Grounding are the metrics designed to evaluate performance in VideoQA and grounded QA, respectively. Other results are taken from~\cite{ExploringQuestionGuidance}.}
	\label{tab:nextgqa-grounding-results}
\end{table*}

\subsection{Video Question Answering with MLLMs}

To perform answer prediction, we employ MLLMs that integrates temporally grounded visual segments and textual queries. Building upon the video-language framework of~\cite{wang2407tarsier, Qwen2.5-VL},  we construct a multimodal input consisting of the original video $\mathcal{V}$,  the natural language question $Q$, and the fused temporal segments $\mathcal{C}_{\mathrm{fused}}$ produced by our grounding module.

To extract representative visual context, we uniformly sample $K_f$ keyframes $\{e_k\}_{k=1}^{K_f}$ from $\mathcal{C}_{\mathrm{fused}}$. Each frame $e_k$ is encoded using the CLIP-ViT~\cite{radford2021clip} visual encoder to obtain patch-level embeddings:
\begin{equation}
\mathbf{E}_v = \text{CLIP-ViT}(\{e_k\}_{i=1}^{K_f}) \in \mathbb{R}^{{K_f} \times N_p \times d_v},
\end{equation}

\noindent where $N_p$ is the number of patches per frame, and $d_v$ denotes the visual embedding dimension.

These features are projected to match the LLM input space via a trainable MLP:
\begin{equation}
\mathbf{H}_v = \text{MLP}(\mathbf{E}_v) \in \mathbb{R}^{K_f \times N_p \times d_h}
\end{equation}
where $d_h$ denotes the hidden dimension required by the language model.

To form the multimodal input, the flattened visual tokens $\mathbf{H}_v^{\text{flat}}$ are concatenated with the embedded textual prompt derived from $Q$ and the candidate options $\mathcal{O} = \{o_i\}_{i=1}^N$ using the answering prompt template $\mathcal{P}_a(Q, \mathcal{O})$:

\begin{gather}
\mathbf{E}_p = \text{Embed}(\mathcal{P}_a(Q, \mathcal{O})) \\
\mathcal{A}_p = \mathcal{M}_a(\text{concat}[\mathbf{W}_p \mathbf{H}_v^{\text{flat}};\ \mathbf{E}_p])
\end{gather}

\noindent where $\mathcal{M}_a$ denotes the MLLM, $\mathbf{W}_p$ is a learnable projection matrix aligning visual features with language embeddings, and the final answer is $\mathcal{A}_p$.

$\mathcal{A}_p$ is generated in free-form text, conditioned on both the temporally grounded visual content and the structured textual prompt. Notably, $\mathcal{P}_a$ differs from the earlier rewriting prompt $\mathcal{P}_r$ used for generating the semantically enriched queries while $\mathcal{P}_r$ is designed to enhance temporal grounding through causal enrichment, $\mathcal{P}_a$ is tailored for answer decoding and decision making based on the fused multimodal evidence.

This final stage enables LeAdQA to jointly reason over visual content and query semantics, completing the VideoQA pipeline with grounded, context-aware answer generation.

\section{Experiments}

\subsection{Experimental Settings}
\subsubsection{Datasets.}
Our evaluation utilizes three established video question answering datasets that collectively assess diverse reasoning capabilities: 

\noindent
\textbf{NExT-QA}~\cite{xiao2021next} is a comprehensive benchmark featuring 5,440 videos (average 44 seconds) with 47,692 multiple-choice questions. Questions are categorized into three reasoning types: temporal action localization (Tem.), causal inference (Cau.), and descriptive analysis (Des.), each with five answer options.

\noindent
\textbf{IntenQA}~\cite{li2023intentqa} focuses on context-aware video intent reasoning with 4,303 videos and 16,297 questions categorized into: Causal Why (CW), Causal How (CH), and temporal action localization (Tem.).

\noindent
\textbf{NExT-GQA}~\cite{xiao2024can} extends NExT-QA by providing visual evidence annotations. It contains 5,417 videos, with 1,557 videos having 10,531 precisely annotated temporal segments corresponding to 8,911 question-answer pairs focused on temporal (Tem.) and causal (Cau.) reasoning.

\subsubsection{Evaluation Metrics.}
For VideoQA, we use accuracy as the main metric. For temporal grounding, we adopt Intersection over Prediction (IoP)~\cite{xiao2024can} and temporal IoU (tIoU) to assess segment containment and overlap with ground truth, reporting both mean scores and hit rates at 0.3 and 0.5 thresholds. We also illustrate Grounded QA Accuracy (Acc@GQA), which considers a prediction correct only if it answers correctly and its temporal segment achieves an $\text{IoP} \geq 0.5$. This reflects a model’s ability to combine semantic understanding with accurate temporal localization.


\subsubsection{Implementation Details.}
Our framework integrates causal reasoning, multimodal alignment, and efficient inference to enable comprehensive video understanding. We first employ GPT-4o~\cite{achiam2023gpt} to enhance question-option pairs by inferring implicit causal relationships via constrained text generation. We adopt UniVTG~\cite{lin2023univtg} as our grounding model, which consists of 4 Transformer encoder layers. Each layer is configured with 1024 hidden dimensions and 8 attention heads, along with specialized output heads for downstream prediction. Each question is equipped with five descriptions and we select top-k ($k \in \{1,3,5\}$) predicted intervals. Multiple overlap thresholds  $ [0.1, 0.3, 0.5, 0.7, 0.9 ]$ guide interval retention or merging decisions based on temporal alignment, enabling visual cue integration. Our final answer generation leverages Tarsier-7B, Tarsier-34B~\cite{wang2407tarsier} and Qwen2.5-VL-3B-Instruct and Qwen2.5-VL-7B-Instruct\cite{Qwen2.5-VL} models with uniform, interval-focused, and hybrid sampling strategies. We evaluate performance across $[1, 2, 4, 8, 16, 32, 48]$ frames to balance accuracy and efficiency. We do our experiments on four A100 40G GPUs.

\subsection{Results and Analysis}

\subsubsection{Baselines.} 
We evaluate LeAdQA on the test sets of \textbf{NExT-GQA} and \textbf{IntentQA}, with the validation set of \textbf{NExT-QA}. 

\noindent For \textbf{NExT-GQA}, we evaluate both multi-choice video QA and temporal grounding tasks, comparing LeAdQA against baselines including IGV~\cite{li2022invariant}, VGT~\cite{xiao2024can}, VIOLETv2~\cite{fu2023empirical}, Temp[Swin], Temp[CLIP], Temp[CLIP(NG+)]~\cite{xiao2024can}, FrozenBiLM~\cite{yang2022zero}, FrozenBiLM (NG+)~\cite{xiao2024can},SeViLA~\cite{yu2024self}, QGAC-TR~\cite{ExploringQuestionGuidance}. We further evaluate Tarsier-7B and Tarsier-34B~\cite{wang2407tarsier}, as well as Qwen2.5VL-3B and Qwen2.5VL-7B~\cite{Qwen2.5-VL}, both with and without LeAdQA integration.

\noindent For \textbf{NExT-QA}, comparisons include video QA methods such as video transformers like InternVideo~\cite{wang2022internvideo}, open-source LLM-based approaches including SeViLA and MVU~\cite{ranasinghe2024understanding}, alongside proprietary LLM-driven models like LLoVi~\cite{zhang2023simple}, VideoAgent~\cite{wang2025videoagent}, MoReVQA~\cite{min2024morevqa}, IG-VLM~\cite{kim2024image}, LangRepo~\cite{kahatapitiya2024language}, LVNet~\cite{park2024lvnet}, and VideoTree~\cite{wang2024videotree}. We also assess Tarsier-7B and Tarsier-34B with and without LeAdQA integration. 

\noindent For \textbf{IntentQA}, we compare LeAdQA with SeViLA, LLoVi, LangRepo, LVNet, and Tarsier models, both standalone and integrated with LeAdQA.

\subsubsection{Comparison with Baselines.}
Table~\ref{tab:nextgqa-grounding-results} presents temporal grounding performance on NExT-GQA, while Tables~\ref{tab:nextgqa videoqa results}, \ref{tab:nextqa results}, and \ref{tab:intentqa results} report VideoQA accuracy on NExT-GQA, NExT-QA, and IntentQA, respectively.

\begin{table}[ht]
    \centering

    \begin{tabular}{lccc}
        \toprule        
        \textbf{Model} & \textbf{Tem.} & \textbf{Cau.} & \textbf{Avg.} \\
        \midrule
        Qwen-3B & 57.4 & 51.4 & 53.9  \\
        \textbf{LeAdQA-3B$'$} & 59.8 (+2.4) & 52.8 (+1.4) & 55.7 (+1.8)  \\
       
        Qwen-7B & 64.4 & 67.3 & 66.1  \\
        \textbf{LeAdQA-7B$'$} & 65.6 (+1.1) & 69.1 (+1.8) & 67.7 (+1.6)  \\
        
        Tarsier-7B & 62.1 & 67.9 & 65.5  \\
        \textbf{LeAdQA-7B} & 63.6 (+1.5) & 70.4 (+2.5) & 67.5 (+2.0)  \\
        Tarsier-34B& \underline{71.7} & \underline{77.7}  & \underline{75.2}  \\
        \textbf{LeAdQA-34B}& \textbf{73.8 (+2.1)} & \textbf{79.4 (+1.7)} & \textbf{77.0 (+1.8)} \\
        \bottomrule
    \end{tabular}
    
    \caption{VideoQA accuracy on NExT-GQA. LeAdQA-3B$'$ and LeAdQA-7B$'$ are developed based on Qwen-3B and Qwen-7B respectively.}
    \label{tab:nextgqa videoqa results}
\end{table}

As demonstrated in Tables~\ref{tab:nextgqa-grounding-results}, \ref{tab:nextgqa videoqa results}, \ref{tab:nextqa results} and \ref{tab:intentqa results}, VideoQA models incorporating visual grounding consistently outperform baseline methods across all datasets. LeAdQA achieves SOTA VideoQA performance by treating temporal grounding as an auxiliary objective that supplements the primary QA task. While existing approaches like SeViLA and VideoTree demonstrate visual localization capabilities, their inability to model causal relationships constrains QA accuracy. LeAdQA addresses this through explicit causal reasoning, proving particularly effective for understanding dynamic processes and event progression. Our results demonstrate that causal reasoning effectively compensates for grounding inaccuracies by providing essential contextual relationships, confirming that visual grounding and temporal alignment are complementary for effective video reasoning.

\begin{table}[ht]
	\centering
	\setlength{\tabcolsep}{1pt} 
\begin{tabular*}{\columnwidth}{@{\extracolsep{\fill}}
		p{4.5cm}  
		>{\centering\arraybackslash}p{0.6cm}  
		>{\centering\arraybackslash}p{0.6cm}  
		>{\centering\arraybackslash}p{0.6cm}  
		>{\centering\arraybackslash}p{0.6cm}  
	}
		\toprule
		\textbf{Model} & \textbf{Tem.} & \textbf{Cau.} & \textbf{Des.} & \textbf{Avg.} \\
		\midrule
		InternVideo~\cite{wang2022internvideo} & 43.4 & 48.0 & 65.1 & 49.1 \\
		SeViLA~\cite{yu2024self} & 61.3 & 61.5 & 75.6 & 63.6 \\
		MVU~\cite{ranasinghe2024understanding} & 55.4 & 48.1 & 64.1 & 55.2 \\
		LLoVi~\cite{zhang2023simple} & 61.0 & 69.5 & 75.6 & 63.6 \\
		VideoAgent~\cite{wang2025videoagent} & 64.5 & 72.7 & 81.1 & 71.3 \\
		MoReVQA~\cite{min2024morevqa} & 56.1 & 52.7 & 71.8 & 60.2 \\
		IG-VLM~\cite{kim2024image} & 63.6 & 69.8 & 74.7 & 68.6 \\
		LangRepo-7B~\cite{kahatapitiya2024language} & 45.7 & 57.8 & 61.9 & 54.6 \\
		LangRepo-12B~\cite{kahatapitiya2024language} & 51.4 & 64.4 & 69.1 & 60.9 \\
		LVNet~\cite{park2024lvnet} & 65.5 & 75.0 & 81.5 & 72.9 \\
		VideoTree~\cite{wang2024videotree} & 67.0 & 75.2 & 81.3 & 73.5 \\
		\midrule
		Tarsier-7B~\cite{wang2407tarsier} & 66.4 & 71.7 & 81.9 & 71.6 \\
		\textbf{LeAdQA-7B} & 66.6 (+0.2) & 72.5 (+0.8) & 82.3 (+0.6) & 72.1 (+0.5) \\
		Tarsier-34B~\cite{wang2407tarsier} & \underline{74.4} & \underline{80.5} & \underline{85.3} & \underline{79.3} \\
		\textbf{LeAdQA-34B} & \textbf{75.7 (+1.3)} & \textbf{81.9 (+1.4)} & \textbf{86.6 (+1.3)} & \textbf{80.6 (+1.3)} \\
		\bottomrule
	\end{tabular*}
	\caption{VideoQA accuracy on NExT-QA.}
	\label{tab:nextqa results}
\end{table}

\begin{table}[ht]
	\centering
	\setlength{\tabcolsep}{3pt}
	\begin{tabular*}{\columnwidth}{@{}
			p{4.2cm} 
			>{\centering\arraybackslash}p{0.8cm} 
			>{\centering\arraybackslash}p{0.8cm} 
			>{\centering\arraybackslash}p{0.8cm} 
			>{\centering\arraybackslash}p{0.8cm} 
			@{}}
		\toprule
		\textbf{Model} & \textbf{CW} & \textbf{CH} & \textbf{Tem.} & \textbf{Avg.} \\
		\midrule
		SeViLA~\cite{yu2024self}                     & -    & -    & -    & 60.9 \\
		LLoVi~\cite{zhang2023simple}                 & 68.4 & 67.4 & 51.1 & 64.0 \\
		IG-VLM~\cite{kim2024image}                   & -    & -    & -    & 64.2 \\
		LangRepo-7B~\cite{kahatapitiya2024language}  & 56.9 & 60.2 & 42.1 & 53.8 \\
		LangRepo-12B~\cite{kahatapitiya2024language} & 62.8 & 62.4 & 47.8 & 59.1 \\
		LVNet~\cite{park2024lvnet}                   & 75.0 & 74.4 & 62.1 & 71.7 \\
		\midrule
		Tarsier-7B                                   & 69.9 & 69.9 & 59.6 & 67.4 \\
		\textbf{LeAdQA-7B}                           & \textbf{71.2} (+1.3) & \textbf{70.2 } (+0.3) & \textbf{60.0 } (+0.4)& \textbf{68.2 } (+0.8)\\
		Tarsier-34B                                  & \underline{79.4} & \underline{78.8} & \underline{69.9} & \underline{76.9} \\
		\textbf{LeAdQA-34B}                          & \textbf{80.4} (+1.0) & \textbf{83.0} (+4.2) & \textbf{70.9} (+1.0) & \textbf{78.5} (+1.6) \\
		\bottomrule
	\end{tabular*}
	\caption{VideoQA accuracy on IntentQA.}
	\label{tab:intentqa results}
\end{table}

Table~\ref{tab:nextgqa videoqa results} demonstrates that our method can utilize various MLLMs as backbones and consistently improve their performance, highlighting the versatility of LeAdQA across different model architectures.
The results in Table~\ref{tab:nextqa results} show that multimodal models extend video understanding capabilities through enhanced semantic alignment and scalability, achieving substantial improvements over LLoVi's caption-based approach. This performance advantage stems from superior cross-modal alignment that overcomes inherent single-modality limitations, particularly in complex reasoning tasks requiring temporal and causal understanding. Table~\ref{tab:intentqa results} reveals consistent performance gains across all question types, with particularly notable improvements in CW (Causal How) questions. This pronounced effect indicates that LLM-based causal reasoning effectively complements visual evidence by reconstructing event chains that conventional approaches typically miss. Notably, the Tarsier-34B model shows greater performance improvements, suggesting that model scale and visual grounding operate synergistically to enhance video comprehension. 


\subsection{Ablation Study}

\subsubsection{Impact of QA Pair Rewriting with GPT-4.}

Table~\ref{tab:abl_grounding_rewriting} demonstrates that GPT-based causal rewriting consistently enhances performance under uniform grounding conditions, with the "+Causal Rewriting" variant achieving superior results across all question categories. The most substantial improvement appears in causal questions (Cau.), validating GPT's proficiency in identifying and modeling causal relationships. Notably, concurrent improvements in temporal and descriptive questions indicate that semantic restructuring strengthens visual-textual alignment beyond the scope of causal reasoning alone.

\begin{table}[ht]
	\centering
	\setlength{\tabcolsep}{3pt}
	\begin{tabular*}{\columnwidth}{@{} p{4.8cm} c c c c @{}}
		\toprule
		\textbf{Setting} & \textbf{Tem.} & \textbf{Cau.} & \textbf{Des.} & \textbf{Avg.} \\
		\midrule
		\multicolumn{5}{l}{\textit{Varying Grounding (w/ Causal Rewriting)}} \\
		\quad Random Sampling       & 72.0 & 79.9 & 84.3 & 78.0 \\
		\quad Uniform Sampling      & 74.4 & 80.5 & 85.3 & 79.3 \\
		\quad Ground-Truth Segments & 79.5 & 82.6 & 85.3 & 82.1 \\
		\midrule
		\multicolumn{5}{l}{\textit{Varying Rewriting (w/ Uniform Grounding)}} \\
		\quad w/o Causal Rewriting        & 74.4 & 80.5 & 85.3 & 79.3 \\
		\quad \textbf{+ Causal Rewriting (ours)} & 75.7 & 81.9 & 86.8 & 80.6 \\
		\bottomrule
	\end{tabular*}
	\caption{Ablation study on temporal grounding and causal rewriting on NExT-QA.}
	\label{tab:abl_grounding_rewriting}
\end{table}

\subsubsection{Impact of Video Temporal Grounding.}
We systematically examine how temporal grounding precision influences answer accuracy on the NExT-QA dataset using Tarsier-34B, maintaining uniform experimental conditions throughout. As illustrated in Table~\ref{tab:abl_grounding_rewriting}, we evaluate three sampling strategies: (1) random frame sampling, (2) uniform keyframe sampling, and (3) ground-truth segment sampling. Our analysis reveals a strong positive correlation between grounding precision and QA performance. The results establish that temporal coherence is fundamental to effective video comprehension, with structured sampling methods significantly outperforming random frame selection. Furthermore, precise visual grounding enhances reasoning quality by directing model attention to relevant visual content, while explicit causal modeling substantially improves understanding of event dynamics, especially in causal reasoning scenarios. These findings collectively underscore that effective video question answering demands the seamless integration of temporal structure, accurate visual localization, and comprehensive causal relationship modeling.


\subsection{In-depth Analysis}

\subsubsection{Parameter Analysis for Interval Fusion.}
Table~\ref{tab:interval fusion} presents our evaluation of the interval fusion strategy on the NExT-QA validation set using Tarsier-34B with uniform 16-frame sampling. We investigate how the number of top-k candidate intervals and the IoU threshold for merging influence overall performance. Our analysis uncovers a critical trade-off in temporal fusion parameters: while increasing K initially improves answer quality by capturing additional visual cues, raising the IoU threshold degrades performance, indicating that overlapping intervals introduce noise that disrupts the reasoning process. An IoU threshold of 0.3 achieves optimal balance, effectively filtering irrelevant temporal segments while preserving essential events.

\begin{table}[ht]
    \setlength{\tabcolsep}{10pt}
    \centering

    \begin{tabular}{lccccc}
        \toprule        
        \multirow{2}{*}{\textbf{Top-K}} & \multicolumn{5}{c}{\textbf{IoU threshold}} \\
        \cmidrule(lr){2-6} 
        ~ & 0.1 & 0.3 & 0.5 & 0.7 & 0.9\\
        \midrule
        \textbf{Top-1} & \textbf{79.0} & 78.9 & 78.9 & 78.8 & 78.6 \\
        \textbf{Top-3} & \textbf{79.5} & \textbf{79.5} & 79.3 & 79.2 & 78.3  \\
        \textbf{Top-5} & \textbf{80.2} & \textbf{80.2} & 79.4 & 79.4 & 73.3  \\
        \bottomrule
    \end{tabular} 
    \caption{Impact of Top-K candidate intervals and IoU thresholds on Accuracy performance in NExT-QA with Tarsier-34B (16 frames).}
    \label{tab:interval fusion}
\end{table}

\subsubsection{Frame Sampling Strategy for Answer Generation.}
\begin{figure}[ht]
    \centering
    \includegraphics[width=1\linewidth]{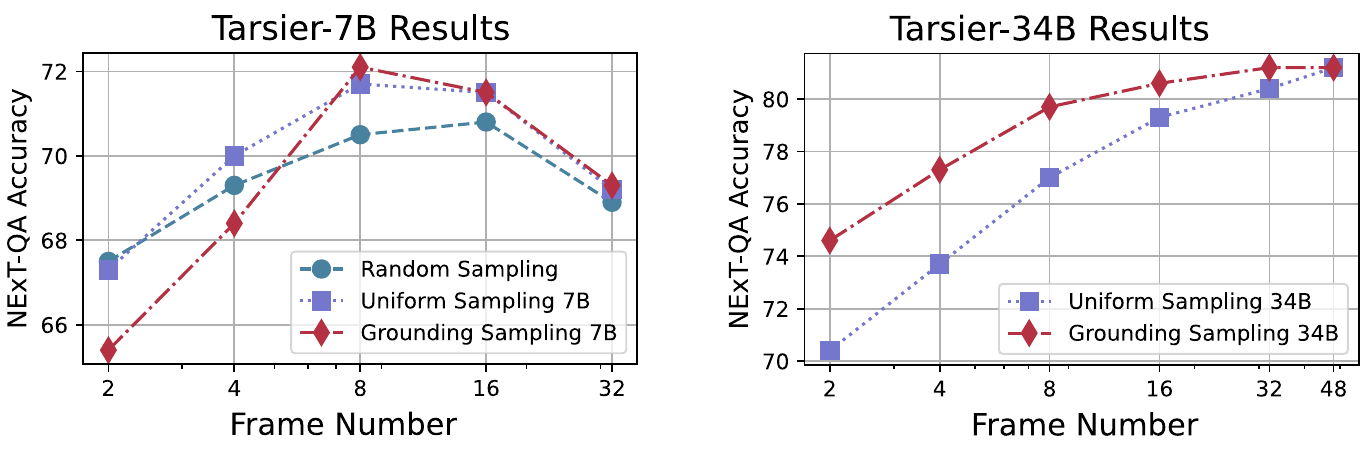}
    \caption{Tarsier-7B (left) and Tarsier-34B (right): VideoQA Accuracy vs. Frame Count.}
    \label{fig:results_7b_with_34b}
\end{figure}

We systematically compare three sampling strategies: random, uniform, and our proposed query-focused sampling within grounded intervals. The results presented in Figure~\ref{fig:results_7b_with_34b}. Our findings demonstrate that random sampling consistently underperforms uniform sampling across all configurations, though temporal sorting narrows this performance gap, reinforcing the significance of temporal coherence. Notably, Figure~\ref{fig:results_7b_with_34b} shows that query-focused sampling achieves comparable accuracy to uniform sampling while using fewer frames (32 vs. 48 frames: 81.2\% vs. 81.2\%), confirming our framework's ability to eliminate irrelevant frames with minimal computational overhead. Furthermore, our experiments reveal distinct optimal frame requirements across model scales: Tarsier-7B reaches peak performance with 8 frames, showing diminishing returns beyond this threshold due to computational constraints, while Tarsier-34B continues improving up to 48 frames while maintaining processing efficiency.

\section{Conclusion}
We present LeAdQA, an efficient framework designed to enhance multimodal reasoning in MLLMs for VideoQA through causal-aware question refinement and context-aware temporal grounding. By reformulating question-option pairs to address causal gaps, LeAdQA facilitates precise grounding of relevant visual content, thereby improving answer accuracy while reducing computational overhead. Experimental results demonstrate consistent improvements in modeling causal relationships and contextual understanding across various video reasoning tasks. Future research will focus on extending this framework to longer videos by exploring hierarchical comprehension and frame token compression techniques for more accurate understanding.

\bibliography{aaai2026}


\end{document}